\documentclass[11pt]{article}

\usepackage[preprint]{acl}

\usepackage{times}
\usepackage{latexsym}
\usepackage[T1]{fontenc}

\usepackage[utf8]{inputenc}

\usepackage{microtype}

\usepackage{inconsolata}

\usepackage{graphicx}
\usepackage{amsmath,hyperref}
\usepackage{amsfonts}
\usepackage{enumitem}
\usepackage[table]{xcolor}
\usepackage{booktabs}
\usepackage{multirow}
\usepackage{stfloats}
\usepackage{makecell}
\usepackage{tabularx,booktabs}
\usepackage{subcaption}
\newcolumntype{Y}{>{\ttfamily\footnotesize\raggedright\arraybackslash}X} 
\newcolumntype{R}{>{\footnotesize\raggedright\arraybackslash}p{2.2cm}} 
\usepackage{xcolor}
\newcommand{\var}[1]{\textcolor{blue!60!black}{\texttt{\{}#1\texttt{\}}}}

\usepackage[most]{tcolorbox}

\newtcolorbox{promptbox}{
  colback=gray!10,    
  colframe=gray!40,   
  boxrule=0.3pt,      
  arc=2pt,            
  left=4pt, right=4pt, top=4pt, bottom=4pt
}

\title{Understanding Structured Financial Data with LLMs: \\A Case Study on Fraud Detection}

\author{Xuwei Tan \thanks{Work done during internship at Coinbase. This paper contains the author’s personal opinions and does not constitute a company policy or statement. These opinions are not endorsed by or affiliated with Coinbase, Inc. or its subsidiaries.}  \\
  The Ohio State University  \\
  Coinbase, Inc.  \\
   \\\And
  Yao Ma \\
  Coinbase, Inc. \\
   \\
   \\\And
  Xueru Zhang \\
  The Ohio State University \\
   \\}

\begin{document}
\maketitle

\begin{abstract}

Detecting fraud in financial transactions typically relies on tabular models that demand heavy feature engineering to handle high-dimensional data and offer limited interpretability, making it difficult for humans to understand predictions. Large Language Models (LLMs), in contrast, can produce human-readable explanations and facilitate feature analysis, potentially reducing the manual workload of fraud analysts and informing system refinements. However, they perform poorly when applied directly to tabular fraud detection due to the difficulty of reasoning over many features, the extreme class imbalance, and the absence of contextual information. To bridge this gap, we introduce FinFRE-RAG, a two-stage approach that applies importance-guided feature reduction to serialize a compact subset of numeric/categorical attributes into natural language and performs retrieval-augmented in-context learning over label-aware, instance-level exemplars. Across four public fraud datasets and three families of open-weight LLMs, FinFRE-RAG substantially improves F1/MCC over direct prompting and is competitive with strong tabular baselines in several settings.  Although these LLMs still lag behind specialized classifiers, they narrow the performance gap and provide interpretable rationales, highlighting their value as assistive tools in fraud analysis.
\end{abstract}

\section{Introduction}
\label{sec:intro}

Financial fraud is a pervasive and evolving threat that costs businesses and consumers billions of dollars every year \cite{hilal2022financial}. The fraud landscape is highly dynamic, with malicious actors continually devising new strategies to circumvent existing safeguards. In response to the limitations of rule-based systems \cite{soui2019rule}, machine learning (ML) models have been adopted to identify illicit or abnormal transactions from large-scale financial data \cite{jin2025analysis}. However, their performance is critically dependent on extensive and laborious feature engineering. This reliance makes them costly to develop and maintain. Furthermore, these models are sensitive to pattern shifts \cite{gama2014survey}, where the statistical properties of the data change over time, requiring frequent retraining and model updates. These trained models also offer limited interpretability beyond aggregate feature importance scores, posing challenges to transparent decision-making.

Recent advances in LLMs offer a promising alternative across many domains \cite{wang-etal-2025-user,hu2024enhancing,ning2025dima,wang2024biorag,tan2026benchmarking}, including finance \cite{wu2023bloomberggpt,wang2023fingpt,yu2024fincon}. With strong capabilities in complex reasoning and emulating the logic of human analysts \cite{brown2020language}, LLMs hold the potential to not only detect fraud but also provide human-readable rationales for their decisions. However, prior studies typically use LLMs as auxiliary tools to augment traditional classifiers rather than as standalone detectors. For example, \cite{yang2025flag} and \cite{huang2025can} integrate LLMs to improve the performance of Graph Neural Networks in fraud detection. Nevertheless, applying LLMs directly to fraud detection remains challenging.  Benchmarks \cite{feng2023empowering,xie2024finben} highlight that existing LLMs still struggle with financial risk prediction, achieving only marginal improvements over random guessing.

A key question arises: How can we adapt LLMs to understand and detect fraud in tabular transaction data? This challenge stems from a core misalignment between the LLMs and the domain realities. Specifically, we have identified two key issues:
\begin{itemize}[leftmargin=*,noitemsep,topsep=0.0pt]
    \item \textbf{Tabular Input Misalignment:} Unlike frauds such as internet scams or phone scams that involve rich text data \cite{yang2025fraud,singh2025advanced}, financial transaction data are primarily numeric and categorical features in a tabular format. LLMs, being pretrained mainly on natural language, may not interpret the semantic meaning of structured features or handle numerical values with the required precision. In addition, real transactions can also involve hundreds or thousands of features, including all of them in a prompt may exceed context limits and introduce noise, as many are less significant signals. This lack of semantic context and high dimensionality makes it difficult for LLMs to detect fraud patterns from raw tabular data.
    
    \item \textbf{Fraud Ambiguity and Rarity:} Fraudulent transactions typically require nuanced, expert knowledge to recognize, and what constitutes “fraud” can vary greatly across different institutions, product types, or regions. An LLM without domain-specific guidance can be easily confounded by this variability in fraud definitions and patterns, leading to unpredictable or nonsensical outputs. Additionally, fraud datasets are usually extremely imbalanced, where the proportion of fraudulent instances may be less than 1\% of all transactions. LLMs may struggle to identify the subtle patterns that differentiate a legitimate transaction from a fraudulent one.
    
\end{itemize}
The key is to teach LLMs what constitutes fraud and which feature patterns often lead to fraudulent behavior. Without explicit guidance, LLMs tend to treat tabular inputs as arbitrary numbers, producing nearly random predictions. To narrow this gap, we leverage selective historical transactions as few-shot exemplars, providing the model with concrete cases of both fraudulent and legitimate behavior. By grounding the task in real examples, the LLM gains a contextual understanding of how subtle feature patterns correspond to risk.

To this end, this paper introduces a \textbf{FIN}ancial \textbf{F}eature-\textbf{RE}duced in-context learning framework with \textbf{R}etrieval-\textbf{A}ugmented \textbf{G}eneration (\textbf{FinFRE-RAG}), designed to harness the reasoning capabilities of LLMs while addressing their inherent weaknesses in structured fraud detection. FinFRE-RAG is built on two key stages: the offline feature reduction for prompt construction with the most impactful features, ensuring compact yet informative representations; and the online retrieval-augmented in-context learning to construct a temporary, task-relevant micro-dataset within the LLM’s context window. Our method recasts fraud detection as an instance-based reasoning problem. By supplying few-shot examples, the LLM adapts to each query without the computational overhead of training. Our contributions are summarized as follows: 
\begin{itemize}[leftmargin=*,nosep]
    \item We propose \textbf{FinFRE-RAG}, a two-stage framework that adapts LLMs to structured fraud detection without task-specific training.
    \item We conduct a systematic evaluation across multiple fraud benchmarks and open-weight LLMs, showing substantial F1/MCC gains bridging prior gaps by demonstrating that, with targeted feature reduction and retrieval, LLMs no longer systematically lag behind traditional methodologies.
    \item We provide principled analyses that clarify how to use LLMs for tabular fraud by quantifying the required number of features and similar transactions, assessing the impact of prediction granularity, and comparing against fine-tuned LLMs.
    
\end{itemize}

\section{Related Work}

\subsection{Fraud Detection}

Traditional approaches for fraud detection relied heavily on rule-based systems and hand-crafted features, which, while interpretable, often struggled to adapt to evolving fraud patterns. In recent years, with the increasing availability of large-scale transaction data and the advances in deep learning, research has shifted towards more expressive models that capture more complex patterns. Traditional machine learning methods, particularly ensemble algorithms such as XGBoost \cite{chen2016xgboost}, LightGBM \cite{ke2017lightgbm}, and CatBoost \cite{prokhorenkova2018catboost}, have shown strong effectiveness in capturing non-linear interactions among structured features. Beyond ensembles, deep neural networks have also been applied to detect different kinds of fraud \cite{fiore2019using, dou2020enhancing, li2024sefraud, yu2023group}.

\subsection{LLMs for finance}  

A growing line of research has emphasized the importance of benchmarking LLMs for financial applications. \citet{xie2023pixiu,xie2024finben} build a large-scale benchmark covering a wide range of financial tasks, including fraud detection.  InvestorBench \cite{li-etal-2025-investorbench} evaluates LLM-based agents in financial decision-making settings, such as stock and cryptocurrency trading.  Beyond text, FinMME \cite{luo-etal-2025-finmme} and FCMR \cite{kim-etal-2025-fcmr} extended the scope to multi-modal evaluation, incorporating tables, charts, and textual reports.

Parallel to benchmark construction, researchers have also introduced a series of domain-specific financial LLMs. Early efforts include FinBERT \cite{yang2020finbert}, which adapted BERT for financial sentiment analysis; BloombergGPT \cite{wu2023bloomberggpt}, a large-scale model trained on proprietary financial corpora; and FinGPT \cite{wang2023fingpt}, which emphasizes open, low-cost, and continuously updated pipelines. Beyond general-purpose models, several frameworks have been designed for task-specific applications. For example, \citet{yu2024fincon} proposed a multi-agent framework for stock trading and portfolio management, while \citet{aguda2024large} explored leveraging LLMs as data annotators to extract relations in financial documents.

\subsection{In-context learning and RAG}

LLMs can adapt to new tasks at inference by conditioning on a few input–label examples in the prompt, without parameter updates \cite{brown2020language}.This \emph{in-context learning} (ICL) enables few-shot generalization through text alone. Prior work suggests that ICL may emulate implicit learning mechanisms, such as pattern matching or regression, while remaining sensitive to the choice and order of demonstrations \cite{min2022rethinking}.

Retrieval-augmented generation (RAG) equips LLMs with external memory: a retriever selects relevant examples, and the generator conditions on them to produce grounded outputs \cite{lewis2020retrieval}. The integration of external knowledge can enhance the accuracy of model responses and improve generation reliability \cite{guu2020retrieval,karpukhin2020dense,ram2023context}. We extend this idea to structured data: retrieve semantically similar instances, present them as context, and let the model reason by comparison.

 \begin{figure*}[htb]
    \centering
    \includegraphics[width=0.95\linewidth]{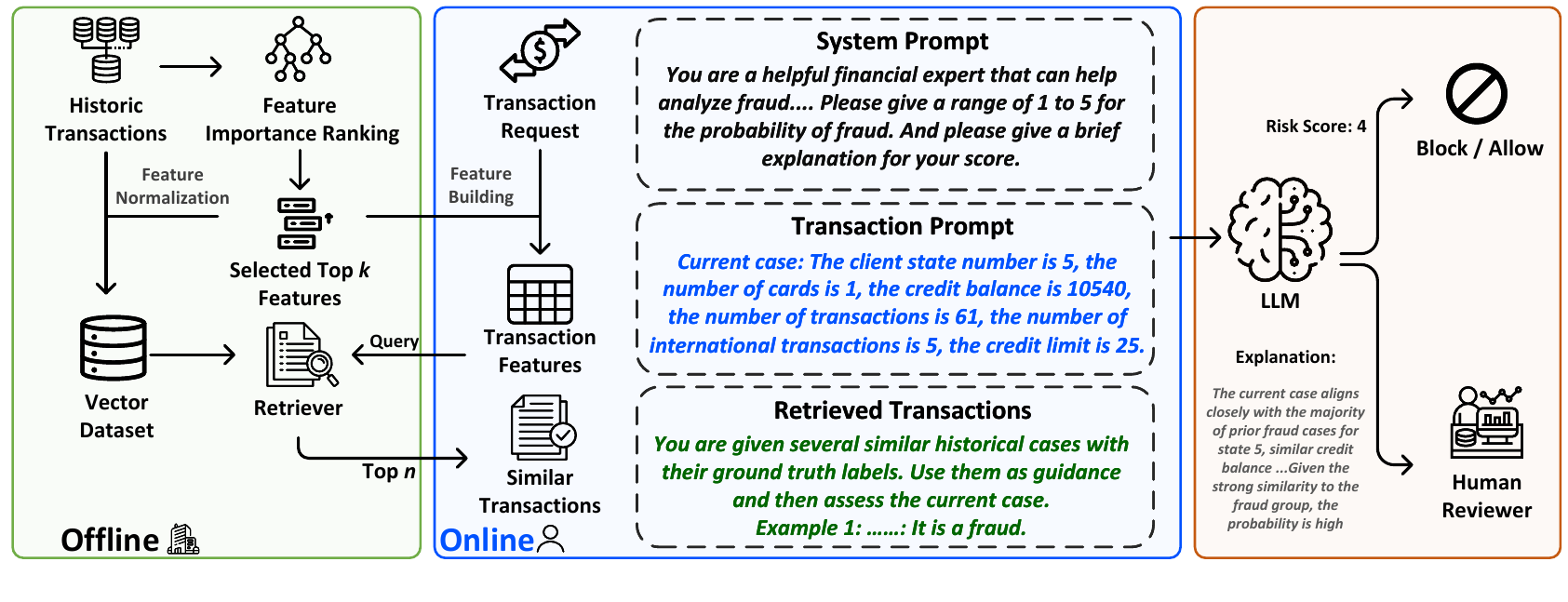}
    \vspace{-12pt}
     \caption{Overall architecture of the FinFRE-RAG framework.}
    \vspace{-12pt}
    \label{fig:frame}
\end{figure*}

\section{Proposed Method}
\label{sec:method}

In this section, we introduce FinFRE-RAG; the overall workflow is illustrated in Figure \ref{fig:frame}. It consists of two key stages: offline feature reduction and online retrieval-augmented in-context learning.

\subsection{Problem Setup}

Let $\mathcal{F}$ denote the set of available features with dimensionality $d = |\mathcal{F}|$. We consider two datasets: (1) \textbf{External historical dataset}:  $\mathcal{D}_{\mathrm{ext}} = \{(x_i, y_i)\}_{i=1}^N$. Each $x_i \in \mathbb{R}^d$ is a transaction represented by the full feature set $\mathcal{F}$. The corresponding label $y_i \in \{0,1\}$ indicates whether the transaction is fraudulent ($y_i = 1$) or legitimate. (2) \textbf{Test dataset}: $\mathcal{D}_{\mathrm{test}} = \{x_q\}_{q=1}^M$. It contains \(M\) unlabeled online query transactions, each \(x_q \in \mathbb{R}^d\)

Our goal is to design a prompting strategy that enables an LLM, denoted by $h_{\theta}(\cdot)$, to classify a query transaction $x_q \in \mathcal{D}_{\mathrm{test}}$. The LLM should achieve this by retrieving relevant examples from $\mathcal{D}_{\mathrm{ext}}$ and providing a risk score with natural language explanations for its decision, without fine-tuning of its parameters $\theta$.

\subsection{Offline Feature Reduction}

Real-world transaction data often includes hundreds or even thousands of features, many of which carry less significant signal for fraud detection. If such high-dimensional data are fed directly into a retrieval-augmented LLM \cite{lewis2020retrieval}, two problems arise: (i) long prompts exceed the model’s context window, making it infeasible to include all features and examples due to extra inference time and memory requirements, and (ii) irrelevant or noisy features dilute the predictive signal and hinder the model’s ability to learn meaningful patterns. Thus, in FinFRE-RAG, we consider reducing the feature space to ensure that the LLM focuses on the most informative attributes and that retrieval remains efficient.

 To address this, we perform an offline feature selection step that leverages the strengths of traditional machine learning models, which are used to capture the initial predictive patterns in structured data. Specifically,  we train a Random Forest on the external dataset $\mathcal{D}_{\mathrm{ext}}$ and extract feature importance scores to rank all $d$ features in $\mathcal{F}$. The goal is not to build a production-grade model, but to obtain a rough yet effective ranking at minimal cost, avoiding the expense of hyperparameter optimization. From this ranking, we retain the top-$k$ features ($k < d$), yielding a reduced feature set $\mathcal{F}_{\mathrm{sel}} \subset \mathcal{F}$.  This reduction ensures that subsequent retrieval-augmented prompts concentrate on the most relevant attributes, allowing the LLM to operate within its context limits.
 
To further align tabular structure with retrieval efficiency, we precompute standardized representations for all numeric features in $\mathcal{F}_{\mathrm{sel}}$. Specifically, for each feature $f \in \mathcal{F}_{\mathrm{sel}}$, we calculate global statistics $(\mu_f, \sigma_f)$ on $\mathcal{D}_{\mathrm{ext}}$ and normalize as:
\begin{equation}
z_f(x) = \frac{x[f] - \mu_f}{\sigma_f}.
\end{equation}
which yields a vectorized dataset where every transaction is embedded in a comparable space. By shifting this computation offline, we remove redundant per-query processing at inference time.

\subsection{Online Retrieval-Augmented In-Context Learning}
\label{sec:onlineretrivel}

At inference, each query transaction $x_q$ is evaluated through a retrieval-augmented reasoning pipeline that casts fraud detection as case-based in-context learning for LLMs. It consists of two stages: instance retrieval and prompt-based reasoning.

\noindent\textbf{1. Instance Retrieval:} We partition the reduced feature set $\mathcal{F}_{\mathrm{sel}}$ into categorical $\mathcal{F}_{\mathrm{cat}}$ and numeric $\mathcal{F}_{\mathrm{num}}$ subsets, where categorical attributes anchor retrieval to structurally similar cases, and numeric attributes enable fine-grained similarity search.

\noindent\underline{Categorical filtering.} 
Let $\mathcal{F}_{\text{cat}}=\{f_{(1)},\dots,f_{(C)}\}$ denote the selected categorical attributes sorted by importance (high to low). We build the candidate pool by progressively adding equality constraints and backing off if the intersection becomes empty:
\begin{equation}
\small
    \mathcal{C}^{(j)} = \{x \in \mathcal{C}^{(j-1)} | x[f_{(j)}] = x_q[f_{(j)}]\},\  j=1,\dots,{C}.
\end{equation}
where $\mathcal{C}^{(0)} = \mathcal{D}_{\text{ext}}$. We stop at the largest $j^\star$ such that $\mathcal{C}^{(j^\star)}\neq\emptyset$ as the final candidate set  $\mathcal{C}(x_q)$.

\noindent \underline{Numeric similarity search.} Within the candidate pool, we retrieve neighbors based on cosine similarity in the feature space (constructed offline):
\begin{equation}
s(x, x_q) = \frac{\langle z(x), z(x_q)\rangle}{\|z(x)\|_2 \cdot \|z(x_q)\|_2}, \ x \in \mathcal{C}(x_q).
\end{equation}
The top-$n$ similar transactions constitute the retrieval set $\mathcal{D}_{\mathrm{retrieved}}(x_q)$. 
The hybrid retrieval strategy combines semantic consistency (via categorical filtering) with graded similarity (via vector search), aligning the selection with the LLM’s strength at analogical reasoning \cite{yasunaga2024large}.

\noindent\textbf{2. Prompt Construction and In-Context Learning:}  
\label{sec:onlineretrieval}
Each retrieved example $x \in \mathcal{D}_{\mathrm{retrieved}}(x_q)$ is converted into a descriptive, human-readable sentence that covers only the features in $\mathcal{F}_{\mathrm{sel}}$, followed by its ground-truth label expressed in natural language (“It is a fraud.” / “It is not a fraud.”). The query case is serialized in the same schema but presented without a label. Rather than listing key–value pairs, we embed all features into a natural-language template. The final prompt provided to the LLM has three components (using the \textsc{ccFraud} dataset prompt as an example):

\begin{promptbox}
\small
\begin{itemize}[leftmargin=*, nosep]
\item  \textbf{Task instruction}: “You are given several similar historical cases with their ground truth labels. Use them as guidance and then assess the current case.”
\item \textbf{Few-shot examples}: “Example 1: The client is a \{\}, the state number is \{\}, the number of cards is \{\},  … It is a fraud...”

\item  \textbf{Query}: “Current case: The client is \{\}, the state number is \{\}, the number of cards is...”
\end{itemize}
\end{promptbox}
The prompt comprises the task instruction, a series of few-shot examples with the query transaction. By contextualizing structured feature values in natural language, we leverage LLMs to interpret tabular data and to draw analogies between the query and retrieved transactions. This formulation reframes fraud detection as case-based reasoning rather than directly classifying tabular inputs.

\section{Experiments}
\label{sec:exp}

We answer the following research questions (RQ):

\begin{itemize}[leftmargin=*, nosep]
    \item \textbf{RQ1}: How do modern LLMs perform on tabular financial fraud detection relative to traditional machine-learning baselines, and does FinFRE-RAG improve their performance?
    \item \textbf{RQ2}: How many features or relevant transactions are needed for LLMs to understand and detect the fraud? 
    \item \textbf{RQ3}: What is the contribution of FinFRE-RAG's feature reduction module to overall performance?
    \item \textbf{RQ4}: How does the granularity of the LLM’s output affect fraud detection performance?
    \item \textbf{RQ5:} Can FinFRE-RAG outperform  LLMs fine-tuned on fraud datasets?
\end{itemize}

\subsection{Experiment Setup}

\noindent\textbf{Datasets. } We evaluate on four public fraud detection datasets: \textsc{ccf} \cite{feng2023empowering}, \textsc{ccFraud} \cite{feng2023empowering, kamaruddin2016credit}, \textsc{IEEE-CIS} \cite{howard2019ieee}, and \textsc{PaySim} \cite{lopez2017synthetic}, covering real-world and synthetic transaction data with varying sizes, feature spaces, and fraud ratios. Detailed dataset descriptions are provided in the Appendix \ref{app:dataset}. To ensure computational feasibility for inference while preserving class imbalance, we randomly sample 8,000 transactions for testing and 2,000 for validation, maintaining the original fraud ratio. The remaining transactions constitute the retrieval pool (external dataset $\mathcal{D}_{\mathrm{ext}}$) during inference.

\noindent\textbf{Models.} Considering the sensitive nature of financial data, which often must remain strictly in-house for compliance, we focus on open-weight instruction-tuned LLMs that can be deployed on-premise. Concretely, we evaluate \textit{Qwen3-14B} and \textit{Qwen3-Next-80B-A3B-Thinking}  \cite{yang2025qwen3}, \textit{Gemma 3-12B} and \textit{Gemma 3-27B} \cite{team2025gemma}, as well as \textit{GPT-OSS-20B} and \textit{GPT-OSS-120B} \cite{agarwal2025gpt}. These models represent a range of medium to large instruction-tuned LLMs. As non-LLM baselines for tabular fraud detection, we include \textit{Random Forest}, \textit{XGBoost}, and the recent deep tabular model \textit{TabM} \cite{gorishniy2024tabm}, covering widely used tree ensembles and a strong state-of-the-art neural baseline.  There are also some domain‑specific financial LLMs, such as FinGPT \cite{yang2023fingpt} and FinMA \cite{xie2023pixiu}. However, these models are based on earlier Llama versions \cite{touvron2023llama2,touvron2023llama} and are comparatively smaller, which may weaker instruction-following and reasoning for our setting, so we do not include them in experiments.

\noindent\textbf{Evaluation Metrics.} Due to the class imbalance, accuracy is not informative. We therefore use F1-score and Matthews Correlation Coefficient (MCC) \cite{chicco2020advantages} as primary metrics, as they better reflect performance under imbalance. We also report precision (share of flagged transactions that are truly fraudulent) and recall (share of frauds correctly identified) for analyst reference.

\noindent\textbf{Implementation Details.} Experiments are run on \mbox{4$\times$A100} GPUs. Results reported are an average of three runs.
Unless otherwise specified, we retain the top \(k=10\) features and retrieve \(n=20\) nearest neighbors to construct the in-context examples (Section~\ref{sec:rq2} analyzes sensitivity to \(k\) and \(n\)). For supervised baselines, we train on each dataset’s \emph{external split} using all features and perform hyperparameter optimization (detailed in Appendix Table~\ref{tab:hparam-spaces-upd}) on the validation set for 50 trials.  Instead of letting LLMs directly output binary predictions, we consider a 5-point risk score, where ${Score}\,\geq\,4$ is viewed as fraud (positive). We present the full prompts we used in Appendix \ref{app:prompts}.

We finetune \textit{Qwen3\mbox{-}14B}, \textit{Gemma~3\mbox{-}12B}, and \textit{GPT\mbox{-}OSS\mbox{-}20B} via LoRA \cite{hu2022lora} using \textsc{unsloth} \cite{unsloth}, on the \emph{external split} only. Since they are tuned on the external split, we do not apply RAG on these models; instead, we directly run inference for each transaction. We conduct a small grid over key hyperparameters on \(10\%\) of the data to select rank, \(\alpha\), learning rate, and learning rate schedulers, then train for one epoch. Full finetuning details (adapter targets, rank, learning rates, and batch size) are in Appendix~\ref{app:hparams}.

\begin{table*}[htb]
\setlength{\tabcolsep}{4pt}
\centering
\resizebox{\textwidth}{!}{
\begin{tabular}{lcccc|cccc|cccc|cccc}
\toprule
\multirow{2}{*}{\textbf{Model}} & \multicolumn{4}{c}{\textbf{\textsc{ccf}}} & \multicolumn{4}{c}{\textbf{\textsc{ccFraud}}} & \multicolumn{4}{c}{\textbf{\textsc{IEEE‑CIS}}} & \multicolumn{4}{c}{\textbf{\textsc{PaySim}}}\\
\cmidrule(lr){2-5}\cmidrule(lr){6-9}\cmidrule(lr){10-13}\cmidrule(lr){14-17}
 & F1 & MCC & Prec. & Rec. & F1 & MCC & Prec. & Rec. & F1 & MCC & Prec. & Rec. & F1 & MCC & Prec. & Rec. \\
\midrule
Qwen3-14B & 0.00 & -0.01 & 0.00 & 0.00 & 0.14 & 0.09 & 0.07 & 0.80 & 0.04 & -0.01 & 0.03 & 0.05 & 0.00 & -0.05 & 0.00 & 0.46  \\
\rowcolor{gray!30}
\ + \textbf{FinFRE-RAG} & 0.31 & 0.36 & 0.20 & 0.64 & 0.48 & 0.46 & 0.59 & 0.41 & 0.62 & 0.60 & 0.61 & 0.63 & 0.11 & 0.22 & 0.06 & 0.85 \\
Qwen3-Next-80B & 0.01 & 0.07 & 0.01 & 0.85 & 0.12 & 0.04 & 0.06 & 0.99 & 0.08 & 0.05 & 0.04 & 0.77 & 0.00 & 0.00 & 0.00 & 1.00  \\
\rowcolor{gray!30}
\ + \textbf{FinFRE-RAG} & 0.08 & 0.19 & 0.04 & 0.86 & 0.48 & 0.45 & 0.43 & 0.55 &  0.35 & 0.38 & 0.23 & 0.75  & 0.09 & 0.19 & 0.05 & 0.85 \\
Gemma 3‑12B & 0.00 & 0.00 & 0.00 & 0.07 & 0.13 & 0.09 & 0.07 & 0.97 & 0.01 & -0.03 & 0.01 & 0.01 & 0.00 & 0.01 & 0.00 & 1.00  \\
\rowcolor{gray!30}
\ + \textbf{FinFRE-RAG} & 0.79 & 0.80 & 0.68 & 0.93 & 0.59 & 0.57 & 0.60 & 0.59 & 0.59 & 0.57 & 0.59 & 0.59  & 0.71 & 0.72 & 0.61 & 0.85 \\
Gemma 3‑27B & 0.00 & 0.01 & 0.00 & 0.43  & 0.12 & 0.05 &  0.06 & 0.99 & 0.03 & -0.02 & 0.02 & 0.05 & 0.00 & 0.01 & 0.00 & 1.00 \\
\rowcolor{gray!30}
\ + \textbf{FinFRE-RAG} & 0.79 & 0.78 & 0.79 & 0.79 & 0.55 & 0.53 & 0.57 & 0.54 & 0.64 & 0.63 & 0.66 & 0.63 & 0.73 & 0.74 & 0.65 & 0.85 \\
GPT‑OSS‑20B & 0.02 & 0.04 & 0.01 & 0.21 & 0.12 & 0.04 & 0.07 & 0.83 & 0.03 & 0.00 & 0.03 & 0.03 & 0.00 & -0.03 & 0.00 & 0.61  \\
\rowcolor{gray!30}
\ + \textbf{FinFRE-RAG} & 0.24 & 0.32 & 0.15 & 0.71 & 0.51 & 0.49 & 0.54 & 0.49 & 0.61 & 0.59 & 0.54 & 0.68 & 0.17 & 0.28 & 0.10 & 0.85  \\
GPT‑OSS‑120B & 0.01 & 0.06 & 0.01 & 0.64 & 0.14 & 0.08 & 0.08 & 0.77 & 0.04 & -0.01 & 0.03 & 0.08 & 0.00 & 0.02 & 0.00 & 1.00 \\
\rowcolor{gray!30}
\ + \textbf{FinFRE-RAG} & 0.44 & 0.46 & 0.33 & 0.64 & 0.53 & 0.51 & 0.63 & 0.46 & 0.66 & 0.64 & 0.64 & 0.68 & 0.25 & 0.36 & 0.15 & 0.85 \\
\midrule

Random Forest & 0.85 &  0.85 & 0.92 & 0.78 & 0.52 & 0.52 & 0.39 & 0.82 & 0.55 & 0.54 & 0.46 & 0.68 & 0.79 & 0.81 & 0.65 & 1.00 \\
XGBoost & 0.89 &  0.89 & 0.92 & 0.86 & 0.48 & 0.50 & 0.32 & 0.92  & 0.74 & 0.73 & 0.67 & 0.82 & 0.68 & 0.71 & 0.55 & 0.92 \\
TabM & 0.85 &  0.85 & 0.92 & 0.79 & 0.66 & 0.65 & 0.71 & 0.62 & 0.82 & 0.82 & 0.88 & 0.76 & 0.92 & 0.92 & 1.00 & 0.85 \\

\bottomrule
\end{tabular}}
\caption{Performance comparison of baseline models and FinFRE‑RAG on fraud datasets.  Each block reports F1, MCC, Precision, and Recall; higher scores indicate better performance.}
    \vspace{-8pt}
\label{tab:mainexp}
\end{table*}

\subsection{Results}
\paragraph{RQ1: Baseline LLM performance and impact of FinFRE‑RAG.}
We present the main comparison results in Table \ref{tab:mainexp}, where we applied the same prompt templates as FinFRE-RAG. The key difference is that baseline LLMs only use the query transaction without retrieval-augmented generation. 

\noindent\textbf{Baseline performance.} The results for baseline LLMs are consistent with findings in \cite{xie2024finben}. LLMs perform poorly when directly prompted to classify fraud data without additional context, and they are substantially below strong tabular methods like \textit{TabM}. For instance, \textit{Qwen3‑14B} achieves a negative MCC score on the \textsc{ccf}, indicating that the model’s predictions on this dataset are indistinguishable from random guessing. Similar trends hold for other models. These results confirm that LLMs, when applied to challenging financial tabular data, fail to capture fraud patterns.

\noindent\textbf{FinFRE‑RAG performance.} Incorporating FinFRE‑RAG with existing LLMs dramatically improves their performance across all datasets. \textit{Qwen3‑14B}’s F1‑scores increase from 0.00 to 0.31 on \textsc{ccf}, from 0.14 to 0.48 on \textsc{ccFraud}, from 0.04 to 0.62 on \textsc{IEEE‑CIS}, and from 0.00 to 0.11 on \textsc{PaySim}. Correspondingly, MCC rises from negative or near‑zero values to 0.36, 0.46, 0.60, and 0.22, respectively. All of the other models benefit from the FinFRE‑RAG and significantly achieve better MCC. Some results are comparable or even exceed the training-based methods like \textit{Random Forest} and \textit{XGBoost}. These gains illustrate that FinFRE‑RAG effectively mitigates the difficulties LLMs face when reasoning over raw tabular inputs. By restricting the prompt to a small set of informative features and providing relevant historical examples, the LLM learns to map numeric and categorical patterns to fraud risk scores.

\paragraph{RQ2: How many features or relevant transactions are needed for LLMs to understand and detect the fraud? }
\label{sec:rq2}

\begin{figure}[ht]
\centering
\begin{minipage}[c]{0.49\columnwidth}
  \centering
  \includegraphics[width=\linewidth,trim={10 10 10 10},clip]{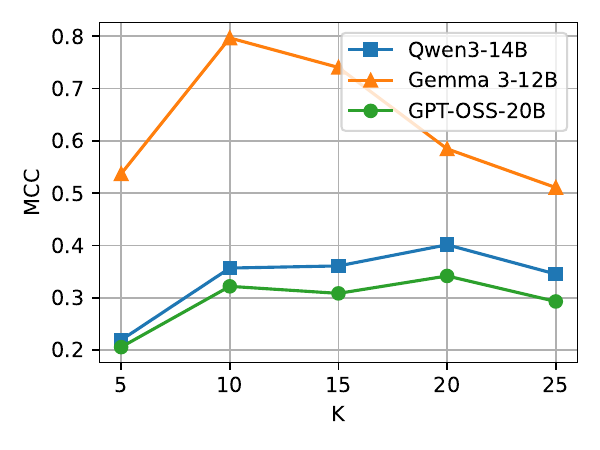}
  \subcaption{\textsc{ccf}}
\end{minipage}
\begin{minipage}[c]{0.49\columnwidth}
  \centering
  \includegraphics[width=\linewidth,trim={10 10 10 10},clip]{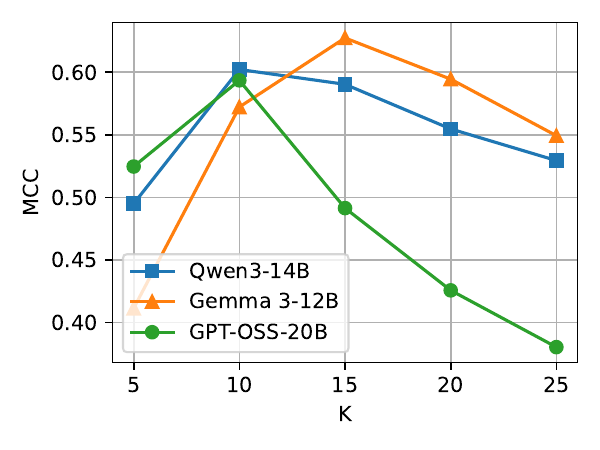}
  \subcaption{\textsc{IEEE‑CIS}}
\end{minipage}
    \vspace{-6pt}
\caption{MCC vs. number of selected features \(k\).}
    \vspace{-6pt}
\label{fig:k-ablation}
\end{figure}

\begin{figure}[t]
\centering
\begin{minipage}[c]{0.49\columnwidth}
  \centering
  \includegraphics[width=\linewidth,trim={10 10 10 10},clip]{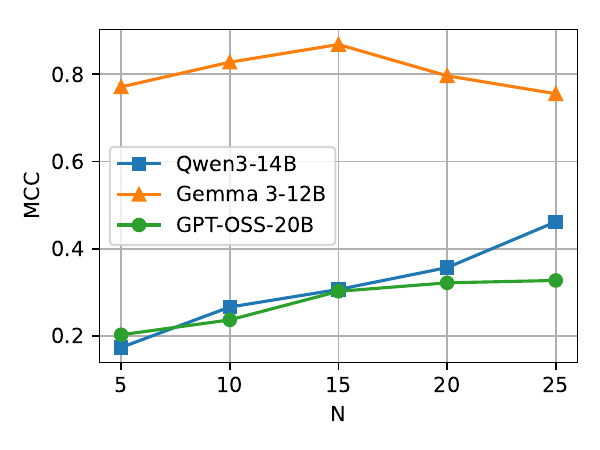}
  \subcaption{\textsc{ccf}}
\end{minipage}
\begin{minipage}[c]{0.49\columnwidth}
  \centering
  \includegraphics[width=\linewidth,trim={10 10 10 10},clip]{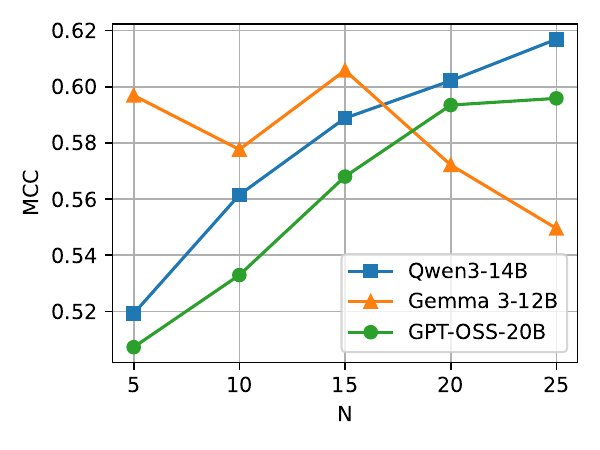}
  \subcaption{\textsc{IEEE‑CIS}}
\end{minipage}
    \vspace{-6pt}
\caption{MCC vs. number of retrieved transactions $n$.}
\label{fig:n-ablation}
    \vspace{-6pt}
\end{figure}

We first conduct experiments to study the impacts of the feature-reduction parameter $k$ and the number of retrieved transactions $n$. 
We vary the number of retained attributes from 5 to 25 and report MCC on the \textsc{ccf} and \textsc{IEEE-CIS} datasets, both of which contain more than 25 features. As shown in Figure~\ref{fig:k-ablation}, simply retaining more features does not necessarily translate into improved performance. Instead, \textbf{the models achieve their best results when provided with a compact yet informative subset of high-impact attributes.} This finding underscores the importance of controlling feature dimensionality to avoid overwhelming the LLM with irrelevant inputs.

We next examine the effect of the number of retrieved transactions $n$ used for in-context examples (Figure~\ref{fig:n-ablation}). Two general trends emerge. First, reasoning models, such as \textit{Qwen3‑14B} and \textit{GPT‑OSS‑20B}, consistently benefit from larger retrieval sets. On \textsc{ IEEE-CIS}, for example, \textit{Qwen3-14B} improves from an MCC of 0.3 to 0.45 as $n$ increases, while \textit{GPT-OSS-20B} rises from 0.34 to 0.43. Similar gains are observed on \textsc{ccf}. These improvements suggest that larger retrieval sets enable reasoning models to extract more nuanced relationships between features with more historical cases. In contrast, \textit{Gemma 3}, as a non-reasoning model, is more sensitive to $n$ and exhibits diminishing gains when too many examples are included. To further investigate \textbf{why additional retrieved samples may degrade \textit{Gemma-3}’s performance}, we analyzed its prediction behavior and found that \textit{Gemma-3} often relies on the aggregate distribution of retrieved examples, particularly the proportion of fraudulent (positive) transactions, as a primary signal for classification. In comparison, \textit{Qwen3-14B} and \textit{GPT-OSS-20B} consider both the number of positive instances and feature-level differences between the query and the retrieved samples, reasoning about how these variations relate to fraud likelihood. Consequently, when the retrieval pool for \textit{Gemma-3} includes more heterogeneous or contradictory examples (e.g., less-similar transactions with opposite labels), its decision confidence declines, leading to poorer performance.

\paragraph{RQ3: Contribution of importance-guided feature reduction.}

\begin{table}[t]
\centering
\resizebox{\linewidth}{!}{
\begin{tabular}{l|cc|cc}
\toprule
 & \multicolumn{2}{c|}{\textbf{Original}} & \multicolumn{2}{c}{\textbf{Random}} \\
\midrule
\textbf{Dataset} & F1 & MCC & F1 & MCC \\
\midrule
\textsc{ccf} & 0.31 / 0.79 / 0.24 & 0.36 / 0.80 / 0.32 & 0.26 / 0.64 / 0.21 & 0.30 / 0.65 / 0.28 \\
\textsc{IEEE-CIS} & 0.62 / 0.59 / 0.61 & 0.60 / 0.57 / 0.59  & 0.27 / 0.31 / 0.34  & 0.32 / 0.37 / 0.32 \\
\bottomrule
\end{tabular}
}
\caption{Feature selection ablation study. Each cell lists the values in order Qwen/Gemma/GPT-OSS.}
    \vspace{-8pt}
\label{tab:feature_selection}
\end{table}

We compare FinFRE-RAG against a variant that retains the same number of attributes chosen uniformly at random. Because \textsc{ccFraud} and \textsc{PaySim} have too few attributes for meaningful ranking, this ablation is still conducted on \textsc{ccf} and \textsc{IEEE-CIS} only. Table \ref{tab:feature_selection} summarizes results for the three models (\textit{Qwen3‑14B}, \textit{Gemma3‑12B}, and \textit{GPT‑OSS‑20B}). In all cases, selecting features by importance consistently outperforms random selection in both F1 and MCC. The effect is especially pronounced on \textsc{IEEE-CIS}. This dataset includes hundreds of numerical and categorical features, making it more likely that random selection captures irrelevant or redundant signals that dilute the predictive context provided to the LLM. By contrast, \textsc{ccf} has only 30 attributes, reducing the risk of severe noise introduction when sampling at random. Consequently, performance degradation is larger on \textsc{IEEE-CIS} than on \textsc{ccf}. These results highlight the value of prioritizing the most informative attributes. By focusing the model’s reasoning on high-signal features, importance-guided reduction enhances the reliability of retrieval-augmented fraud detection, ensuring that the LLM is not overwhelmed by noisy or extraneous inputs.

\paragraph{RQ4: Comparing the granularity of LLM’s output. }
Prior studies typically frame LLM‑based fraud detection as a direct binary classification task: the model is asked to decide whether a transaction is “bad” or “good”. In this work, we instead ask the LLM to produce a 5‑point risk score: Score 1 indicating the lowest probability of fraud and Score 5 the highest, along with a brief explanation, for a better interpretable generation to human analysts. However, it is not obvious whether increasing output granularity enhances or hinders detection accuracy.  To answer this question, we compare these two prompting strategies under the same setting. Table~\ref{tab:scoring_vs_binary} presents F1 and MCC across four datasets.

\begin{table}[t]
\centering
\resizebox{\linewidth}{!}{
\begin{tabular}{l|cc|cc}
\toprule
 & \multicolumn{2}{c|}{\textbf{Scoring}} & \multicolumn{2}{c}{\textbf{Binary}}\\
\midrule
\textbf{Dataset} & F1 & MCC & F1 & MCC\\
\midrule  
\textsc{ccf} & 0.31 / \textbf{0.79 / 0.24} & 0.36 / \textbf{0.80 / 0.32} & \textbf{0.49} / 0.26 / 0.10 & \textbf{0.51} / 0.37 / 0.18\\
\textsc{ccFraud} & 0.48 / \textbf{0.59 / 0.51}  & 0.46 / \textbf{0.57 / 0.49} & \textbf{0.5} / 0.53 / 0.40 & \textbf{0.47} / 0.51 / 0.37\\
\textsc{IEEE‑CIS} & \textbf{0.62 / 0.59 / 0.61} & \textbf{0.60 / 0.57 / 0.59} & 0.51 / 0.33 / 0.42 & 0.51 / 0.36 / 0.40 \\
\textsc{PaySim} &\textbf{ 0.11 / 0.71 / 0.17} &\textbf{ 0.22 / 0.72 / 0.28} &  0.07 / 0.26 / 0.15 & 0.17 / 0.34 / 0.26\\

\bottomrule
\end{tabular}
}
\caption{Risk scoring vs. binary output. Each cell lists the values in order Qwen/Gemma/GPT-OSS.}
    \vspace{-12pt}
\label{tab:scoring_vs_binary}
\end{table}

 \begin{figure*}[htb]
    \centering
    \includegraphics[width=1\linewidth]{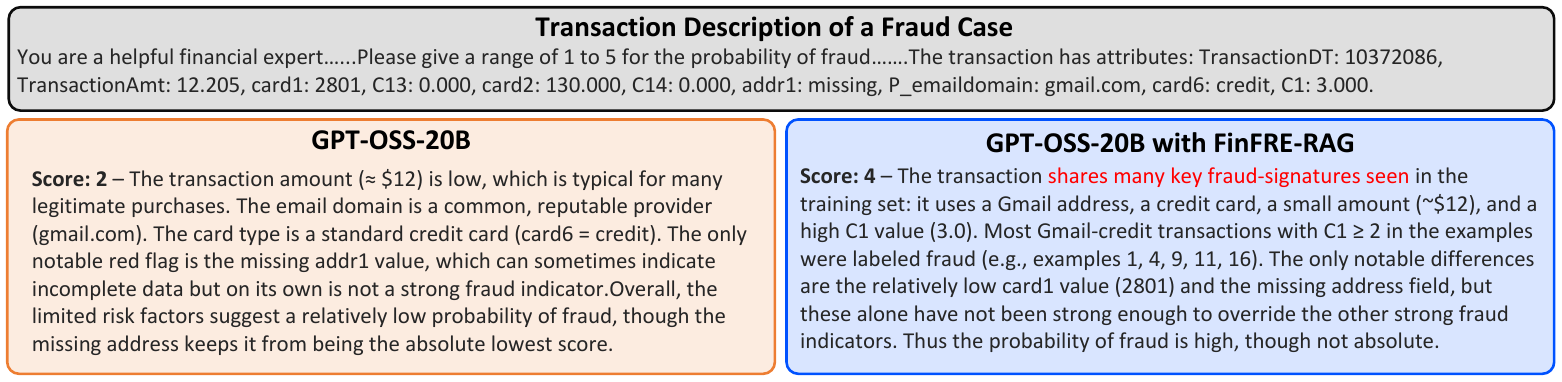}
    \vspace{-10pt}
     \caption{Example responses from GPT-OSS-20B without and with FinFRE-RAG. With in-context learning on similar transactions, it learns to use patterns between different transactions to identify potential fraud.}
    \vspace{-8pt}
    \label{fig:response}
\end{figure*}

Our results demonstrate that fine-grained scoring yields benefits across most datasets and models. Converting the decision into a 5-level risk score improves both F1 and MCC relative to a binary response in most cases. These results suggest that risk scoring allows models to leverage uncertainty more effectively, extracting additional signal from ambiguous cases that would otherwise be forced into a hard binary decision. Overall, we find that risk scoring not only yields more informative output but also improves detection performance. 

\paragraph{RQ5: Can fine-tuning LLMs yield better detections?}
We adopted two prompting regimes: \emph{descriptive} and \emph{schema-grounded} (Appendix~\ref{app:prompts}) to fine-tune three LLMs. Concretely, we use two different types of datasets, the simulated (\textsc{ccFraud}) and the real-world (\textsc{IEEE-CIS}), with \emph{descriptive} and \emph{schema-grounded} prompts, respectively. Models are fine-tuned to predict fraud labels from the formatted inputs. Since these datasets contain only binary labels, to keep training–evaluation consistent, we restrict outputs to binary classification without explanations.

\begin{table}[htb]
\setlength{\tabcolsep}{6pt}
\centering
\resizebox{\columnwidth}{!}{
\begin{tabular}{lcccc|cccc}
\toprule
\multirow{2}{*}{\textbf{Model}} & \multicolumn{4}{c}{\textbf{\textsc{ccFraud}}} & \multicolumn{4}{c}{\textbf{\textsc{IEEE-CIS}}} \\
\cmidrule(lr){2-5}\cmidrule(lr){6-9}
 & F1 & MCC & Prec. & Rec. & F1 & MCC & Prec. & Rec. \\
\midrule
\textbf{Qwen3-14B w/ ours}   & 0.48 & 0.46 & 0.59 & 0.41 & 0.62 & 0.60 & 0.61 & 0.63 \\
\textbf{Qwen3-14B FT}   & 0.55 & 0.52 & 0.62 & 0.49 & 0.52 & 0.53 & 0.71 & 0.41  \\
\textbf{Gemma 3-12B w/ ours} & 0.59 & 0.57 & 0.60 & 0.59 & 0.59 & 0.57 & 0.59 & 0.59 \\
\textbf{Gemma 3-12B FT} & 0.57 & 0.55 & 0.48 & 0.69 &  0.40 & 0.44 & 0.72 &0.28 \\
\textbf{GPT-OSS-20B w/ ours} & 0.51 & 0.49 & 0.54 & 0.49 & 0.61 & 0.59 & 0.54 & 0.68 \\
\textbf{GPT-OSS-20B FT} & 0.61 & 0.58 & 0.62 & 0.60 & 0.45 & 0.47 & 0.68 & 0.34 \\
\bottomrule
\end{tabular}}
\caption{Finetuned LLMs (FT) vs. FinFRE-RAG (ours). }
    \vspace{-8pt}
\label{tab:finetuneexp}
\end{table}

Direct fine-tuning on task data does \emph{not} consistently outperform FinFRE-RAG. Across our settings, FT models tend to exhibit {higher precision but lower recall} than FinFRE-RAG, indicating potential under-coverage of diverse fraud modes and/or overfitting to the majority class in imbalanced data. By contrast, FinFRE-RAG maintains a more balanced precision–recall profile, likely because retrieval exposes the model to heterogeneous, label-aware exemplars at inference. These observations underscore that instance-grounded reasoning, rather than parameter updates alone, is a key driver of robustness under class imbalance. Moreover, because tabular datasets contain limited natural language, pure fine-tuning biases models toward narrow binary prediction heads, diminishing their general generative capacity and precluding case-specific explanations. At the current stage, ICL with retrieval offers a more practical pathway than fine-tuning for financial fraud detection. Advancing domain-specific fine-tuning methods that preserve reasoning and explanation on structural data remains an important direction for future work.

\subsection{Further analysis and discussion}

\noindent \textbf{Behavioral differences.} To illustrate the behavioral differences induced by our framework, we present \textit{GPT-OSS-20B}’s responses with and without FinFRE-RAG on a fraudulent transaction. When directly making the prediction (Figure \ref{fig:response} left), the model relies predominantly on broad, pretrained priors and surface correlations (e.g., treating the email domain as a direct cue), yielding judgments that are weakly grounded in the tabular evidence. In contrast, with FinFRE-RAG the model conditions on retrieved, label-aware exemplars and an importance-guided subset of attributes, and its rationale explicitly reflects cross-feature interactions. Concretely, FinFRE-RAG response calibrates risk by aligning the query against historical cases and by weighing feature-level deltas, producing explanations that mirror dataset-specific fraud regularities rather than spurious textual heuristics.

\noindent \textbf{Why Gemma 3 outperforms others significantly on \textsc{ccf} and \textsc{PaySim}?} As we stated in RQ2 it relies on the label distribution within the retrieved neighbors. On \textsc{PaySim} (synthetic) and \textsc{ccf} (with PCA-compressed features), cosine similarity over the feature set yields high-purity neighborhoods: retrieved examples are close and mostly share the same label. Gemma 3 finds a shortcut, leaning on a simple majority signal in the neighborhood to produce high-recall predictions. By contrast, reasoning models sometimes “over-interpret” benign within-cluster differences, which can slightly lower F1 under such clustered conditions. Therefore, in the real-world IEEE-CIS, where neighborhoods are more mixed (heterogeneous labels, cross-feature confounds), we find \textit{Gemma 3} cannot outperform other models without finer cross-reasoning.

\noindent\textbf{\textsc{PaySim} recall uniformity} (bias from label homogeneous neighborhoods). The test split of \textsc{PaySim} contains only 13 positive cases; for 11 of them the retrieved neighbors were all positive, and for the remaining 2 they were all negative. Because the LLMs often rely on the majority label within the retrieved set, these label‑homogeneous neighborhoods produce near‑deterministic outcomes and explain the uniform recall in Table \ref{tab:mainexp}. This exposes a weakness of our method: when neighborhoods are label‑biased, predictions can be skewed by retrieval rather than feature‑level evidence.

\section{Conclusion}
\label{sec:conclusion}
We introduced FinFRE‑RAG, a two‑stage framework, converting selected numerical and categorical attributes into descriptive natural language and grounding each prediction in similar historical cases, to adapt LLMs for fraud detection in tabular financial transactions. FinFRE‑RAG substantially improves LLM performance over direct prompting, making it a promising direction for future research and production. Based on the framework, we further examine key factors impacting LLMs’ understanding of fraud. Overall, FinFRE-RAG provides a practical path toward reliable, transparent fraud assessments and opens avenues for future research.

\clearpage
\newpage
\section*{Limitations}

Our study has several limitations that suggest caution and opportunities for future work. 
\begin{itemize}[leftmargin=*, nosep]
    \item \textbf{Model and baseline scope.} We evaluate a representative set of open-weight LLMs and common tabular baselines, but do not compare against production-grade systems with large feature stores, domain-specialized LLMs, or very large models.
    \item \textbf{Datasets and validity.} For compliance reasons, we rely exclusively on public datasets. Only \textsc{IEEE-CIS} contains a rich set of features; others have relatively few and function more like “toy” datasets (e.g., \textsc{ccFraud}, \textsc{PaySim}). This under-represents real-world settings where volume and feature cardinality are much higher, patterns are subtler, and fraud is harder to capture. Moreover, \textsc{IEEE-CIS} and \textsc{ccf} anonymize feature names; in practice, revealing semantically meaningful feature names to the LLM could further improve reasoning, which we are not able to evaluate here.
    \item \textbf{Metrics and systems considerations.} Our evaluation emphasizes standard classification metrics (F1/MCC). Because the study is research-oriented and our system is not production-tuned, we did not optimize for or report latency, throughput, or serving cost under realistic traffic.
    \item \textbf{Explanations vs. causality.} While we analyze model rationales qualitatively, these are heuristic and conditioned on retrieved exemplars; they should not be interpreted as causal explanations.
    \item \textbf{Retrieval design.} We use cosine similarity over importance-guided features, for example retrieval. Alternative distance metrics, learned retrievers, or more advanced RAG pipelines could yield stronger neighborhoods and further gains.
    \item \textbf{Human evaluation.} Beyond a few illustrative case studies, we lack systematic human reviews. Pattern attributions produced by our pipeline require additional validation by human reviewers.
    \item \textbf{Domain breadth.} This work is a case study in \emph{financial fraud}. Whether our method can be transferred to other domains remains to be tested.
    \item \textbf{Fine-tuning budget.} We employ LoRA-style parameter-efficient fine-tuning and do not fine-tune larger models. This choice may cap achievable performance and limit conclusions.
\end{itemize}

\section*{Ethical Considerations}

\begin{itemize}[leftmargin=*, nosep]
\item \textbf{Data privacy and security.} Fraud datasets may contain PII and sensitive attributes. We recommend strictly on-premise deployment with open-weight models whenever feasible, minimizing data exfiltration risk. 
Retrieval stores should be privacy-hardened to prevent sensitive attributes from leaking through prompts or rationales.

\item \textbf{Transparency and accountability.} FinFRE-RAG produces concise rationales, but these are model-generated and may omit relevant factors. Institutions should treat them as \emph{decision support}, not final determinations. Deployed systems should maintain model cards, data provenance records, retrieval policies, and reviewer notes, and provide users with appropriate notices/explanations for adverse actions. Human review should be mandatory for high-impact outcomes.

\item \textbf{Misuse and overreliance.} Because LLM outputs can appear authoritative, there is a risk of over-trust. We caution against fully autonomous blocking and recommend conservative thresholds, analyst-in-the-loop workflows, and defense-in-depth with orthogonal detectors. Systems should be stress-tested for prompt injection and retrieval poisoning, protected by robust access and usage controls, and monitored for distributional shifts that could degrade performance or fairness over time.
\end{itemize}

\section*{Acknowledgements}
This work was funded in part by the National Science Foundation under award numbers IIS-2202699, IIS-2416895, IIS-2301599, CMMI-2301601, and DMS-2529302.

\bibliography{refs}

\begin{thebibliography}{52}
\providecommand{\natexlab}[1]{#1}

\bibitem[{Agarwal et~al.(2025)Agarwal, Ahmad, Ai, Altman, Applebaum, Arbus, Arora, Bai, Baker, Bao et~al.}]{agarwal2025gpt}
Sandhini Agarwal, Lama Ahmad, Jason Ai, Sam Altman, Andy Applebaum, Edwin Arbus, Rahul~K Arora, Yu~Bai, Bowen Baker, Haiming Bao, and 1 others. 2025.
\newblock gpt-oss-120b \& gpt-oss-20b model card.
\newblock \emph{arXiv preprint arXiv:2508.10925}.

\bibitem[{Aguda et~al.(2024)Aguda, Siddagangappa, Kochkina, Kaur, Wang, and Smiley}]{aguda2024large}
Toyin~D Aguda, Suchetha Siddagangappa, Elena Kochkina, Simerjot Kaur, Dongsheng Wang, and Charese Smiley. 2024.
\newblock Large language models as financial data annotators: A study on effectiveness and efficiency.
\newblock In \emph{Proceedings of the 2024 Joint International Conference on Computational Linguistics, Language Resources and Evaluation (LREC-COLING 2024)}, pages 10124--10145.

\bibitem[{Akiba et~al.(2019)Akiba, Sano, Yanase, Ohta, and Koyama}]{akiba2019optuna}
Takuya Akiba, Shotaro Sano, Toshihiko Yanase, Takeru Ohta, and Masanori Koyama. 2019.
\newblock Optuna: A next-generation hyperparameter optimization framework.
\newblock In \emph{Proceedings of the 25th ACM SIGKDD international conference on knowledge discovery \& data mining}, pages 2623--2631.

\bibitem[{Brown et~al.(2020)Brown, Mann, Ryder, Subbiah, Kaplan, Dhariwal, Neelakantan, Shyam, Sastry, Askell et~al.}]{brown2020language}
Tom Brown, Benjamin Mann, Nick Ryder, Melanie Subbiah, Jared~D Kaplan, Prafulla Dhariwal, Arvind Neelakantan, Pranav Shyam, Girish Sastry, Amanda Askell, and 1 others. 2020.
\newblock Language models are few-shot learners.
\newblock \emph{Advances in neural information processing systems}, 33:1877--1901.

\bibitem[{Chen and Guestrin(2016)}]{chen2016xgboost}
Tianqi Chen and Carlos Guestrin. 2016.
\newblock Xgboost: A scalable tree boosting system.
\newblock In \emph{Proceedings of the 22nd acm sigkdd international conference on knowledge discovery and data mining}, pages 785--794.

\bibitem[{Chicco and Jurman(2020)}]{chicco2020advantages}
Davide Chicco and Giuseppe Jurman. 2020.
\newblock The advantages of the matthews correlation coefficient (mcc) over f1 score and accuracy in binary classification evaluation.
\newblock \emph{BMC genomics}, 21(1):6.

\bibitem[{Daniel~Han and team(2023)}]{unsloth}
Michael~Han Daniel~Han and Unsloth team. 2023.
\newblock \href {http://github.com/unslothai/unsloth} {Unsloth}.

\bibitem[{Dou et~al.(2020)Dou, Liu, Sun, Deng, Peng, and Yu}]{dou2020enhancing}
Yingtong Dou, Zhiwei Liu, Li~Sun, Yutong Deng, Hao Peng, and Philip~S Yu. 2020.
\newblock Enhancing graph neural network-based fraud detectors against camouflaged fraudsters.
\newblock In \emph{Proceedings of the 29th ACM international conference on information \& knowledge management}, pages 315--324.

\bibitem[{Feng et~al.(2023)Feng, Dai, Huang, Zhang, Xie, Han, Chen, Lopez-Lira, and Wang}]{feng2023empowering}
Duanyu Feng, Yongfu Dai, Jimin Huang, Yifang Zhang, Qianqian Xie, Weiguang Han, Zhengyu Chen, Alejandro Lopez-Lira, and Hao Wang. 2023.
\newblock Empowering many, biasing a few: Generalist credit scoring through large language models.
\newblock \emph{arXiv preprint arXiv:2310.00566}.

\bibitem[{Fiore et~al.(2019)Fiore, De~Santis, Perla, Zanetti, and Palmieri}]{fiore2019using}
Ugo Fiore, Alfredo De~Santis, Francesca Perla, Paolo Zanetti, and Francesco Palmieri. 2019.
\newblock Using generative adversarial networks for improving classification effectiveness in credit card fraud detection.
\newblock \emph{Information Sciences}, 479:448--455.

\bibitem[{Gama et~al.(2014)Gama, {\v{Z}}liobait{\.e}, Bifet, Pechenizkiy, and Bouchachia}]{gama2014survey}
Jo{\~a}o Gama, Indr{\.e} {\v{Z}}liobait{\.e}, Albert Bifet, Mykola Pechenizkiy, and Abdelhamid Bouchachia. 2014.
\newblock A survey on concept drift adaptation.
\newblock \emph{ACM computing surveys (CSUR)}, 46(4):1--37.

\bibitem[{Gorishniy et~al.(2024)Gorishniy, Kotelnikov, and Babenko}]{gorishniy2024tabm}
Yury Gorishniy, Akim Kotelnikov, and Artem Babenko. 2024.
\newblock Tabm: Advancing tabular deep learning with parameter-efficient ensembling.
\newblock \emph{arXiv preprint arXiv:2410.24210}.

\bibitem[{Guu et~al.(2020)Guu, Lee, Tung, Pasupat, and Chang}]{guu2020retrieval}
Kelvin Guu, Kenton Lee, Zora Tung, Panupong Pasupat, and Mingwei Chang. 2020.
\newblock Retrieval augmented language model pre-training.
\newblock In \emph{International conference on machine learning}, pages 3929--3938. PMLR.

\bibitem[{Hilal et~al.(2022)Hilal, Gadsden, and Yawney}]{hilal2022financial}
Waleed Hilal, S~Andrew Gadsden, and John Yawney. 2022.
\newblock Financial fraud: a review of anomaly detection techniques and recent advances.
\newblock \emph{Expert systems With applications}, 193:116429.

\bibitem[{Howard and Bouchon-Meunier(2019)}]{howard2019ieee}
Addison Howard and Bernadette Bouchon-Meunier. 2019.
\newblock Ieee cis, inversion, john lei, lynn@ vesta, marcus2010, hussein abbass.
\newblock \emph{IEEE-CIS Fraud Detection, Kaggle}.

\bibitem[{Hu et~al.(2022)Hu, yelong shen, Wallis, Allen-Zhu, Li, Wang, Wang, and Chen}]{hu2022lora}
Edward~J Hu, yelong shen, Phillip Wallis, Zeyuan Allen-Zhu, Yuanzhi Li, Shean Wang, Lu~Wang, and Weizhu Chen. 2022.
\newblock \href {https://openreview.net/forum?id=nZeVKeeFYf9} {Lo{RA}: Low-rank adaptation of large language models}.
\newblock In \emph{International Conference on Learning Representations}.

\bibitem[{Hu et~al.(2024)Hu, Xia, Zhang, Fu, Wu, Huan, Li, Tang, and Zhou}]{hu2024enhancing}
Jun Hu, Wenwen Xia, Xiaolu Zhang, Chilin Fu, Weichang Wu, Zhaoxin Huan, Ang Li, Zuoli Tang, and Jun Zhou. 2024.
\newblock Enhancing sequential recommendation via llm-based semantic embedding learning.
\newblock In \emph{Companion Proceedings of the ACM Web Conference 2024}, pages 103--111.

\bibitem[{Huang and Wang(2025)}]{huang2025can}
Tairan Huang and Yili Wang. 2025.
\newblock Can llms find fraudsters? multi-level llm enhanced graph fraud detection.
\newblock \emph{arXiv preprint arXiv:2507.11997}.

\bibitem[{Jin and Zhang(2025)}]{jin2025analysis}
Jing Jin and Yongqing Zhang. 2025.
\newblock The analysis of fraud detection in financial market under machine learning.
\newblock \emph{Scientific Reports}, 15(1):29959.

\bibitem[{Kamaruddin and Ravi(2016)}]{kamaruddin2016credit}
SK~Kamaruddin and Vadlamani Ravi. 2016.
\newblock Credit card fraud detection using big data analytics: use of psoaann based one-class classification.
\newblock In \emph{Proceedings of the international conference on informatics and analytics}, pages 1--8.

\bibitem[{Karpukhin et~al.(2020)Karpukhin, Oguz, Min, Lewis, Wu, Edunov, Chen, and Yih}]{karpukhin2020dense}
Vladimir Karpukhin, Barlas Oguz, Sewon Min, Patrick~SH Lewis, Ledell Wu, Sergey Edunov, Danqi Chen, and Wen-tau Yih. 2020.
\newblock Dense passage retrieval for open-domain question answering.
\newblock In \emph{EMNLP (1)}, pages 6769--6781.

\bibitem[{Ke et~al.(2017)Ke, Meng, Finley, Wang, Chen, Ma, Ye, and Liu}]{ke2017lightgbm}
Guolin Ke, Qi~Meng, Thomas Finley, Taifeng Wang, Wei Chen, Weidong Ma, Qiwei Ye, and Tie-Yan Liu. 2017.
\newblock Lightgbm: A highly efficient gradient boosting decision tree.
\newblock \emph{Advances in neural information processing systems}, 30.

\bibitem[{Kim et~al.(2025)Kim, Kim, and Kim}]{kim-etal-2025-fcmr}
Seunghee Kim, Changhyeon Kim, and Taeuk Kim. 2025.
\newblock \href {https://doi.org/10.18653/v1/2025.acl-long.1138} {Fcmr: Robust evaluation of financial cross-modal multi-hop reasoning}.
\newblock In \emph{Proceedings of the 63rd Annual Meeting of the Association for Computational Linguistics (Volume 1: Long Papers)}, pages 23352--23380, Vienna, Austria. Association for Computational Linguistics.

\bibitem[{Lewis et~al.(2020)Lewis, Perez, Piktus, Petroni, Karpukhin, Goyal, K{\"u}ttler, Lewis, Yih, Rockt{\"a}schel et~al.}]{lewis2020retrieval}
Patrick Lewis, Ethan Perez, Aleksandra Piktus, Fabio Petroni, Vladimir Karpukhin, Naman Goyal, Heinrich K{\"u}ttler, Mike Lewis, Wen-tau Yih, Tim Rockt{\"a}schel, and 1 others. 2020.
\newblock Retrieval-augmented generation for knowledge-intensive nlp tasks.
\newblock \emph{Advances in neural information processing systems}, 33:9459--9474.

\bibitem[{Li et~al.(2025)Li, Cao, Yu, Javaji, Deng, He, Jiang, Zhu, Subbalakshmi, Huang, Qian, Peng, Suchow, and Xie}]{li-etal-2025-investorbench}
Haohang Li, Yupeng Cao, Yangyang Yu, Shashidhar~Reddy Javaji, Zhiyang Deng, Yueru He, Yuechen Jiang, Zining Zhu, K.p. Subbalakshmi, Jimin Huang, Lingfei Qian, Xueqing Peng, Jordan~W. Suchow, and Qianqian Xie. 2025.
\newblock \href {https://doi.org/10.18653/v1/2025.acl-long.126} {Investorbench: A benchmark for financial decision-making tasks with llm-based agent}.
\newblock In \emph{Proceedings of the 63rd Annual Meeting of the Association for Computational Linguistics (Volume 1: Long Papers)}, pages 2509--2525, Vienna, Austria. Association for Computational Linguistics.

\bibitem[{Li et~al.(2024)Li, Yang, Zhou, Meng, Wang, Wu, Tan, Song, Pan, Yu et~al.}]{li2024sefraud}
Kaidi Li, Tianmeng Yang, Min Zhou, Jiahao Meng, Shendi Wang, Yihui Wu, Boshuai Tan, Hu~Song, Lujia Pan, Fan Yu, and 1 others. 2024.
\newblock Sefraud: Graph-based self-explainable fraud detection via interpretative mask learning.
\newblock In \emph{Proceedings of the 30th ACM SIGKDD Conference on Knowledge Discovery and Data Mining}, pages 5329--5338.

\bibitem[{Lopez-Rojas(2017)}]{lopez2017synthetic}
E~Lopez-Rojas. 2017.
\newblock Synthetic financial datasets for fraud detection.
\newblock \emph{Kaggle. Available online: https://www. kaggle. com/datasets/ealaxi/paysim1 (accessed on 29 July 2023)}.

\bibitem[{Luo et~al.(2025)Luo, Kou, Yang, Luo, Huang, Xiao, Peng, Liu, Ji, Liu, Han, Zhang, and Guo}]{luo-etal-2025-finmme}
Junyu Luo, Zhizhuo Kou, Liming Yang, Xiao Luo, Jinsheng Huang, Zhiping Xiao, Jingshu Peng, Chengzhong Liu, Jiaming Ji, Xuanzhe Liu, Sirui Han, Ming Zhang, and Yike Guo. 2025.
\newblock \href {https://doi.org/10.18653/v1/2025.acl-long.1426} {Finmme: Benchmark dataset for financial multi-modal reasoning evaluation}.
\newblock In \emph{Proceedings of the 63rd Annual Meeting of the Association for Computational Linguistics (Volume 1: Long Papers)}, pages 29465--29489, Vienna, Austria. Association for Computational Linguistics.

\bibitem[{Min et~al.(2022)Min, Lyu, Holtzman, Artetxe, Lewis, Hajishirzi, and Zettlemoyer}]{min2022rethinking}
Sewon Min, Xinxi Lyu, Ari Holtzman, Mikel Artetxe, Mike Lewis, Hannaneh Hajishirzi, and Luke Zettlemoyer. 2022.
\newblock Rethinking the role of demonstrations: What makes in-context learning work?
\newblock \emph{arXiv preprint arXiv:2202.12837}.

\bibitem[{Ning et~al.(2025)Ning, Cai, Li, Fang, Tan, Chai, and Liu}]{ning2025dima}
Yansong Ning, Shuowei Cai, Wei Li, Jun Fang, Naiqiang Tan, Hua Chai, and Hao Liu. 2025.
\newblock Dima: An llm-powered ride-hailing assistant at didi.
\newblock In \emph{Proceedings of the 31st ACM SIGKDD Conference on Knowledge Discovery and Data Mining V. 2}, pages 4728--4739.

\bibitem[{Prokhorenkova et~al.(2018)Prokhorenkova, Gusev, Vorobev, Dorogush, and Gulin}]{prokhorenkova2018catboost}
Liudmila Prokhorenkova, Gleb Gusev, Aleksandr Vorobev, Anna~Veronika Dorogush, and Andrey Gulin. 2018.
\newblock Catboost: unbiased boosting with categorical features.
\newblock \emph{Advances in neural information processing systems}, 31.

\bibitem[{Ram et~al.(2023)Ram, Levine, Dalmedigos, Muhlgay, Shashua, Leyton-Brown, and Shoham}]{ram2023context}
Ori Ram, Yoav Levine, Itay Dalmedigos, Dor Muhlgay, Amnon Shashua, Kevin Leyton-Brown, and Yoav Shoham. 2023.
\newblock In-context retrieval-augmented language models.
\newblock \emph{Transactions of the Association for Computational Linguistics}, 11:1316--1331.

\bibitem[{Singh et~al.(2025)Singh, Singh, and Singh}]{singh2025advanced}
Gurjot Singh, Prabhjot Singh, and Maninder Singh. 2025.
\newblock Advanced real-time fraud detection using rag-based llms.
\newblock \emph{arXiv preprint arXiv:2501.15290}.

\bibitem[{Soui et~al.(2019)Soui, Gasmi, Smiti, and Gh{\'e}dira}]{soui2019rule}
Makram Soui, Ines Gasmi, Salima Smiti, and Khaled Gh{\'e}dira. 2019.
\newblock Rule-based credit risk assessment model using multi-objective evolutionary algorithms.
\newblock \emph{Expert systems with applications}, 126:144--157.

\bibitem[{Tan et~al.(2026)Tan, Hu, and Zhang}]{tan2026benchmarking}
Xuwei Tan, Ziyu Hu, and Xueru Zhang. 2026.
\newblock \href {https://openreview.net/forum?id=GLPmZhhCAE} {Benchmarking bias mitigation toward fairness without harm from vision to {LVLM}s}.
\newblock In \emph{The Fourteenth International Conference on Learning Representations}.

\bibitem[{Team et~al.(2025)Team, Kamath, Ferret, Pathak, Vieillard, Merhej, Perrin, Matejovicova, Ram{\'e}, Rivi{\`e}re et~al.}]{team2025gemma}
Gemma Team, Aishwarya Kamath, Johan Ferret, Shreya Pathak, Nino Vieillard, Ramona Merhej, Sarah Perrin, Tatiana Matejovicova, Alexandre Ram{\'e}, Morgane Rivi{\`e}re, and 1 others. 2025.
\newblock Gemma 3 technical report.
\newblock \emph{arXiv preprint arXiv:2503.19786}.

\bibitem[{Touvron et~al.(2023{\natexlab{a}})Touvron, Lavril, Izacard, Martinet, Lachaux, Lacroix, Rozi{\`e}re, Goyal, Hambro, Azhar et~al.}]{touvron2023llama}
Hugo Touvron, Thibaut Lavril, Gautier Izacard, Xavier Martinet, Marie-Anne Lachaux, Timoth{\'e}e Lacroix, Baptiste Rozi{\`e}re, Naman Goyal, Eric Hambro, Faisal Azhar, and 1 others. 2023{\natexlab{a}}.
\newblock Llama: Open and efficient foundation language models.
\newblock \emph{arXiv preprint arXiv:2302.13971}.

\bibitem[{Touvron et~al.(2023{\natexlab{b}})Touvron, Martin, Stone, Albert, Almahairi, Babaei, Bashlykov, Batra, Bhargava, Bhosale et~al.}]{touvron2023llama2}
Hugo Touvron, Louis Martin, Kevin Stone, Peter Albert, Amjad Almahairi, Yasmine Babaei, Nikolay Bashlykov, Soumya Batra, Prajjwal Bhargava, Shruti Bhosale, and 1 others. 2023{\natexlab{b}}.
\newblock Llama 2: Open foundation and fine-tuned chat models.
\newblock \emph{arXiv preprint arXiv:2307.09288}.

\bibitem[{Wang et~al.(2024)Wang, Long, Xiao, Cai, Wu, Meng, Wang, and Zhou}]{wang2024biorag}
Chengrui Wang, Qingqing Long, Meng Xiao, Xunxin Cai, Chengjun Wu, Zhen Meng, Xuezhi Wang, and Yuanchun Zhou. 2024.
\newblock Biorag: A rag-llm framework for biological question reasoning.
\newblock \emph{arXiv preprint arXiv:2408.01107}.

\bibitem[{Wang et~al.(2025)Wang, Liu, Sun, Ma, Wang, Ma, Su, Chen, Gao, Dalal et~al.}]{wang-etal-2025-user}
Jianling Wang, Yifan Liu, Yinghao Sun, Xuejian Ma, Yueqi Wang, He~Ma, Zhengyang Su, Minmin Chen, Mingyan Gao, Onkar Dalal, and 1 others. 2025.
\newblock User feedback alignment for {LLM}-powered exploration in large-scale recommendation systems.
\newblock In \emph{Proceedings of the 63rd Annual Meeting of the Association for Computational Linguistics (Volume 6: Industry Track)}, pages 996--1003.

\bibitem[{Wang et~al.()Wang, Yang, and Wang}]{wang2023fingpt}
Neng Wang, Hongyang Yang, and Christina Wang.
\newblock Fingpt: Instruction tuning benchmark for open-source large language models in financial datasets.
\newblock In \emph{NeurIPS 2023 Workshop on Instruction Tuning and Instruction Following}.

\bibitem[{Wu et~al.(2023)Wu, Irsoy, Lu, Dabravolski, Dredze, Gehrmann, Kambadur, Rosenberg, and Mann}]{wu2023bloomberggpt}
Shijie Wu, Ozan Irsoy, Steven Lu, Vadim Dabravolski, Mark Dredze, Sebastian Gehrmann, Prabhanjan Kambadur, David Rosenberg, and Gideon Mann. 2023.
\newblock Bloomberggpt: A large language model for finance.
\newblock \emph{arXiv preprint arXiv:2303.17564}.

\bibitem[{Xie et~al.(2024)Xie, Han, Chen, Xiang, Zhang, He, Xiao, Li, Dai, Feng et~al.}]{xie2024finben}
Qianqian Xie, Weiguang Han, Zhengyu Chen, Ruoyu Xiang, Xiao Zhang, Yueru He, Mengxi Xiao, Dong Li, Yongfu Dai, Duanyu Feng, and 1 others. 2024.
\newblock Finben: A holistic financial benchmark for large language models.
\newblock \emph{Advances in Neural Information Processing Systems}, 37:95716--95743.

\bibitem[{Xie et~al.(2023)Xie, Han, Zhang, Lai, Peng, Lopez-Lira, and Huang}]{xie2023pixiu}
Qianqian Xie, Weiguang Han, Xiao Zhang, Yanzhao Lai, Min Peng, Alejandro Lopez-Lira, and Jimin Huang. 2023.
\newblock Pixiu: A comprehensive benchmark, instruction dataset and large language model for finance.
\newblock \emph{Advances in Neural Information Processing Systems}, 36:33469--33484.

\bibitem[{Yang et~al.(2025{\natexlab{a}})Yang, Li, Yang, Zhang, Hui, Zheng, Yu, Gao, Huang, Lv et~al.}]{yang2025qwen3}
An~Yang, Anfeng Li, Baosong Yang, Beichen Zhang, Binyuan Hui, Bo~Zheng, Bowen Yu, Chang Gao, Chengen Huang, Chenxu Lv, and 1 others. 2025{\natexlab{a}}.
\newblock Qwen3 technical report.
\newblock \emph{arXiv preprint arXiv:2505.09388}.

\bibitem[{Yang et~al.(2025{\natexlab{b}})Yang, Liu, Wang, Zhang, Yang, and Shi}]{yang2025flag}
Chengdong Yang, Hongrui Liu, Daixin Wang, Zhiqiang Zhang, Cheng Yang, and Chuan Shi. 2025{\natexlab{b}}.
\newblock Flag: Fraud detection with llm-enhanced graph neural network.
\newblock In \emph{Proceedings of the 31st ACM SIGKDD Conference on Knowledge Discovery and Data Mining V. 2}, pages 5150--5160.

\bibitem[{Yang et~al.(2023)Yang, Liu, and Wang}]{yang2023fingpt}
Hongyang Yang, Xiao-Yang Liu, and Christina~Dan Wang. 2023.
\newblock Fingpt: Open-source financial large language models.
\newblock \emph{FinLLM Symposium at IJCAI 2023}.

\bibitem[{Yang et~al.(2025{\natexlab{c}})Yang, Zhu, Wu, Wang, Yao, Wu, Hu, Li, Wong, and Wang}]{yang2025fraud}
Shu Yang, Shenzhe Zhu, Zeyu Wu, Keyu Wang, Junchi Yao, Junchao Wu, Lijie Hu, Mengdi Li, Derek~F Wong, and Di~Wang. 2025{\natexlab{c}}.
\newblock Fraud-r1: A multi-round benchmark for assessing the robustness of llm against augmented fraud and phishing inducements.
\newblock \emph{arXiv preprint arXiv:2502.12904}.

\bibitem[{Yang et~al.(2020)Yang, UY, and Huang}]{yang2020finbert}
Yi~Yang, Mark Christopher~Siy UY, and Allen Huang. 2020.
\newblock \href {https://arxiv.org/abs/2006.08097} {Finbert: A pretrained language model for financial communications}.
\newblock \emph{Preprint}, arXiv:2006.08097.

\bibitem[{Yasunaga et~al.(2024)Yasunaga, Chen, Li, Pasupat, Leskovec, Liang, Chi, and Zhou}]{yasunaga2024large}
Michihiro Yasunaga, Xinyun Chen, Yujia Li, Panupong Pasupat, Jure Leskovec, Percy Liang, Ed~H. Chi, and Denny Zhou. 2024.
\newblock \href {https://openreview.net/forum?id=AgDICX1h50} {Large language models as analogical reasoners}.
\newblock In \emph{The Twelfth International Conference on Learning Representations}.

\bibitem[{Yu et~al.(2023)Yu, Wang, Wang, Li, Qin, Zhang, Liao, and Zhang}]{yu2023group}
Jianke Yu, Hanchen Wang, Xiaoyang Wang, Zhao Li, Lu~Qin, Wenjie Zhang, Jian Liao, and Ying Zhang. 2023.
\newblock Group-based fraud detection network on e-commerce platforms.
\newblock In \emph{Proceedings of the 29th ACM SIGKDD conference on knowledge discovery and data mining}, pages 5463--5475.

\bibitem[{Yu et~al.(2024)Yu, Yao, Li, Deng, Jiang, Cao, Chen, Suchow, Cui, Liu et~al.}]{yu2024fincon}
Yangyang Yu, Zhiyuan Yao, Haohang Li, Zhiyang Deng, Yuechen Jiang, Yupeng Cao, Zhi Chen, Jordan Suchow, Zhenyu Cui, Rong Liu, and 1 others. 2024.
\newblock Fincon: A synthesized llm multi-agent system with conceptual verbal reinforcement for enhanced financial decision making.
\newblock \emph{Advances in Neural Information Processing Systems}, 37:137010--137045.

\end{thebibliography}

\clearpage
\newpage
\appendix

\section{Datasets}
\label{app:dataset}

\begin{table*}[ht]
\centering
\resizebox{\linewidth}{!}{
\begin{tabular}{lrrrrrr}
\toprule
\textbf{Dataset} & 
\makecell{\textbf{\# Numerical}\\\textbf{Features}} & 
\makecell{\textbf{\# Categorical}\\\textbf{Features}} & 
\makecell{\textbf{\# Positive}} & 
\makecell{\textbf{\# Negative}} & 
\makecell{\textbf{Imbalanced}\\(\%)}& 
\makecell{\textbf{License}} \\
\midrule
\textbf{\textsc{ccf}}        & 30 & - & 490 & 284{,}317 & 0.172 & DbCL v1.0 \\
\textbf{\textsc{ccFraud}}  & 5 & 2 & 596,014 & 9,403,986 & 5.98 & DbCL v1.0 \\
\textbf{\textsc{IEEE-CIS}}  & 373 & 20 & 20,663 & 569,877 & 3.50 & Competition Data  \\
\textbf{\textsc{PaySim}}  & 5 & 3 & 8,213 & 6,354,407 & 0.13 & CC BY-SA 4.0 \\
\bottomrule
\end{tabular}
    }
\caption{Dataset summary. “\# ” denotes feature counts; dashes indicate not applicable. }
\label{tab:data_summary}
\end{table*}

\noindent We use four publicly available fraud detection datasets (stats shown in Table \ref{tab:data_summary}) in our experiments. Below, we provide detailed descriptions:

\begin{itemize}[leftmargin=*]

    \item \textbf{\textsc{ccf} \footnote{https://www.kaggle.com/datasets/mlg-ulb/creditcardfraud}} \cite{feng2023empowering}: A credit card fraud dataset with 284{,}807 transactions from European cardholders in September 2013. Each record includes 30 features, 28 of which are anonymized via PCA for confidentiality. The dataset is highly imbalanced, with only 0.172\% fraudulent cases.

    \item \textbf{\textsc{ccFraud} \footnote{https://github.com/The-FinAI/CALM/blob/main/data/fraud detection/ccFraud/}} \cite{feng2023empowering, kamaruddin2016credit}: It is a \textbf{simulated} dataset contains about 1 million transactions with 7 features (e.g., gender, account balance, transaction count). The proportion of fraudulent samples is 5.98\%.

    \item \textbf{\textsc{IEEE-CIS} \footnote{https://www.kaggle.com/competitions/ieee-fraud-detection/overview}} \cite{howard2019ieee}: IEEE-CIS is a \textbf{real-world} dataset. We use the transaction-level data, which consists of 590{,}540 transactions with hundreds of anonymized numerical and categorical features. Fraudulent transactions make up 3.5\% of the data, presenting challenges in both dimensionality and class imbalance. The license can be found on Kaggle \footnote{https://www.kaggle.com/competitions/ieee-fraud-detection/rules}.

    \item \textbf{\textsc{PaySim} \footnote{https://www.kaggle.com/datasets/ealaxi/paysim1}} \cite{lopez2017synthetic}: A large-scale \textbf{simulated} mobile money transaction dataset with 6.3 million samples. Each transaction is labeled as fraud or non-fraud, with features such as transaction type, amount, and balance updates. The fraud ratio is 0.13\%. As noted in the dataset documentation, some columns (\texttt{oldbalanceOrg}, \texttt{newbalanceOrig}, \texttt{oldbalanceDest}, \texttt{newbalanceDest}) can lead to information leakage, as models may exploit them as shortcuts for fraud detection. We nevertheless include these features, since removing them would leave very limited information for learning. Therefore, evaluation results on PaySim should be interpreted with caution, and more emphasis should be placed on the other three datasets.

\end{itemize}

\noindent To ensure computational feasibility for LLM inference while preserving class imbalance, we randomly sample 8,000 transactions from each dataset for testing and 2,000 for validation, maintaining the original fraud ratio. The remaining transactions constitute the retrieval pool used by FinFRE-RAG during inference. Fraudulent transactions are treated as positive samples in all evaluations. For missing values, we mark them as "missing" in LLM inferences.

\section{Hyperparameters}
\label{app:hparams}

\paragraph{LLM inference.} For LLM inference, we set the LLM temperature to 0.6 to balance determinism and diversity in rationales and nucleus sampling to 0.95. The max length is set to 16,384.  We retain the top \(k=10\) features for datasets and retrieve \(n=20\) nearest neighbors to construct the in-context examples. For those datasets with fewer than $10$ features, we use all features during inference.

\paragraph{Baseline Training.} We tune all supervised baselines with Optuna~\cite{akiba2019optuna}, using \textbf{50 trials} per dataset–model pair.
Each trial is trained on $\mathcal{D}_\mathrm{ext}$; the configuration with the best validation MCC is selected and evaluated on the test set. For TabM training, we deployed early stopping with patience of 5 epochs. For tree methods, we set class weights via \texttt{scale\_pos\_weight} (XGBoost) or \texttt{class\_weight=balanced} (RF) to mitigate the impact of class imbalance.
Table~\ref{tab:hparam-spaces-upd} lists the search spaces.

\paragraph{LLM Fine-tuning.} We apply parameter-efficient finetuning (PEFT) with LoRA \cite{hu2022lora} to \textit{Qwen3-14B}, \textit{Gemma~3-12B}, and \textit{GPT-OSS-20B} using Unsloth \cite{unsloth}. 
To control computation cost, we conduct a \emph{small grid search} on a {10\%} subsample of the external split ($\mathcal{D}_{ext}$) and select the configuration with the {best validation loss}. The selected configuration is then trained on $D_{ext}$ for one epoch. Our grid varies only the learning rate, scheduler, and LoRA rank; all other settings are fixed for stability and comparability. We always use 10\% iterations for the warm-up.

\begin{table*}[htbp]
\centering
\begin{tabular}{llp{8.8cm}}
\toprule
\textbf{Model} & \textbf{Hyperparameter} & \textbf{Search Space (Optuna)} \\
\midrule
\multirow{9}{*}{Random Forest} 
& n\_estimators & \texttt{suggest\_int}(50, 200) \\
& max\_depth & \texttt{suggest\_int}(5, 20) \\
& min\_samples\_split & \texttt{suggest\_int}(2, 10) \\
& min\_samples\_leaf & \texttt{suggest\_int}(1, 5) \\
& max\_features & \texttt{suggest\_categorical}([\texttt{``sqrt''}, \texttt{``log2''}, \texttt{None}]) \\
& bootstrap & \texttt{suggest\_categorical}([\texttt{True}, \texttt{False}]) \\
& class\_weight & fixed to \texttt{``balanced''} \\
\midrule
\multirow{9}{*}{XGBoost} 
& n\_estimators & \texttt{suggest\_int}(50, 200) \\
& max\_depth & \texttt{suggest\_int}(3, 10) \\
& learning\_rate & \texttt{suggest\_float}(0.01, 0.5) \\
& subsample & \texttt{suggest\_float}(0.6, 1.0) \\
& colsample\_bytree & \texttt{suggest\_float}(0.6, 1.0) \\
& reg\_alpha & \texttt{suggest\_float}(0, 10) \\
& reg\_lambda & \texttt{suggest\_float}(0, 10) \\
& scale\_pos\_weight & fixed to $r=\frac{\#\text{negative}}{\#\text{positive}}$ (training split) \\
\midrule
\multirow{8}{*}{TabM} 
& n\_blocks & \texttt{suggest\_int}(2, 4) \\
& d\_block & \texttt{suggest\_categorical}([128, 256, 512]) \\
& k & \texttt{suggest\_categorical}([16, 32, 64]) \\
& dropout & \texttt{suggest\_float}(0.0, 0.3) \\
& learning\_rate & \texttt{suggest\_float}(1e{-}4, 1e{-}2, log=True) \\
& embedding\_method & fixed to \texttt{``piecewise linear embedding''} \\
& n\_bins & \texttt{suggest\_int}(16, 48) \\
& d\_embedding & \texttt{suggest\_categorical}([4, 8, 16, 32]) \\
\bottomrule
\end{tabular}
\caption{Optuna search spaces used in our experiments for all baselines.}
\label{tab:hparam-spaces-upd}
\end{table*}

\begin{table*}[htbp]
\centering
\begin{tabular}{lp{10cm}}
\toprule
\textbf{Dimension} & \textbf{Values} \\
\midrule
Learning rate & \{2e$-$4, 5e$-$5, 1e$-$5, 5e$-$6\} \\
Scheduler & \{linear, cosine\} \\
LoRA rank ($r$) & \{16, 32, 64\} \\
\midrule
\multirow{8}{*}{Fixed settings} & Precision: bfloat16 \\ 
& Optimizer: AdamW (8-bit), weight decay $=10^{-2}$ \\ 
& LoRA $\alpha=2 \times r$, LoRA dropout $=0.05$ \\
& Targeted modules: \texttt{q\_proj}, \texttt{k\_proj}, \texttt{v\_proj}, \texttt{o\_proj}, \texttt{gate\_proj}, \texttt{up\_proj}, \texttt{down\_proj} \\
& Seq length $=2048$ \\ 
& Batching: \texttt{batch size=8}, \texttt{gradient accumulation steps=2}\\
& Eval steps $=1000$ \\
\bottomrule
\end{tabular}
\caption{Tuning grid and fixed training settings for LLMs fine-tuning.}
\label{tab:peft-grid-small}
\end{table*}

\section{Prompts}
\label{app:prompts}
We provide the full prompts in Table~\ref{app:tab:prompts}, including the system prompt and dataset-specific prompts. Some datasets have features with anonymized or symbolically named fields (e.g., \texttt{V1--V28}) that carry limited semantic content for language models. For these datasets (\textsc{ccf} and \textsc{IEEE-CIS}), we adopt \textit{schema-grounded prompts}, which present the raw feature names and values directly.  For datasets with semantically interpretable attributes (e.g., \textsc{ccFraud}, \textsc{PaySim}), we employ \textit{descriptive prompts}, which paraphrase feature names into concise natural language (e.g., ``the transaction amount is \texttt{\$\{TransactionAmt\}}'' rather than ``\texttt{C1=...}''). For retrieved exemplars, we format each exemplar with the same schema-grounded or descriptive convention as the target instance to avoid distribution shift. At inference time, the model is instructed to output a calibrated fraud score (Score: 1--5) and a brief rationale.

\begin{table*}[htbp]
\centering
\begin{tabularx}{\linewidth}{@{} R Y @{}}
\toprule
\textbf{Role / Dataset} & \textbf{Prompt} \\
\midrule
System & You are a helpful financial expert that can help analyze fraud. Use the fewest reasoning steps needed to reach a correct answer. Please give a score of 1 to 5 for the probability of fraud. You must include \textbf{Score:} in your response. For example, \textbf{Score: 1} means the lowest probability of fraud, and \textbf{Score: 5} means the highest probability of fraud. Provide a brief explanation for your score. \\
\hline
RAG    & \begin{minipage}[t]{\linewidth}\ttfamily\footnotesize
You are given several similar historical cases with their ground truth labels.\\
Use them as guidance and then assess the current case.\\[2pt]
Example 1: \var{example\_transaction\_prompt} It is a fraud/It is not a fraud.\\
\ldots\\
Example N: \var{example\_transaction\_prompt} It is a fraud/It is not a fraud.
\end{minipage}  \\
\hline
ccf & \begin{minipage}[t]{\linewidth}\footnotesize
The client has only numerical input variables which are the result of a PCA transformation: 
V10: \var{V10}, V14: \var{V14}, V4: \var{V4}, V12: \var{V12}, V11: \var{V11}, 
V17: \var{V17}, V3: \var{V3}, V7: \var{V7}, V16: \var{V16}, V2: \var{V2}.
\end{minipage} \\
\hline
ccFraud      & \begin{minipage}[t]{\linewidth}\footnotesize
The client is a \var{gender}. the state number is \var{state}, the number of cards is \var{cardholder}, the credit balance is \var{balance},
the number of transactions is \var{numTrans}, the number of international transactions is \var{numIntlTrans},
the credit limit is \var{creditLine}.
\end{minipage} \\
\hline
IEEE-CIS & \begin{minipage}[t]{\linewidth}\footnotesize
The transaction has attributes: 
TransactionDT: \var{TransactionDT}, TransactionAmt: \var{TransactionAmt}, card1: \var{card1}, C13: \var{C13}, 
card2: \var{card2}, C14: \var{C14}, addr1: \var{addr1}, P\_emaildomain: \var{P\_emaildomain}, 
card6: \var{card6}, C1: \var{C1}.
\end{minipage} \\
\hline
PaySim      & \begin{minipage}[t]{\linewidth}\footnotesize
The transaction has: step is \var{step}, transaction type is \var{type}, amount is \var{amount}, originator is \var{nameOrig}, original balance before transaction is \var{oldbalanceOrg}, originator balance after transaction is \var{newbalanceOrig}, recipient is \var{nameDest}, recipient balance before transaction is \var{oldbalanceDest}, recipient balance after transaction is \var{newbalanceDest}.
\end{minipage} \\
\bottomrule
\end{tabularx}
\caption{Prompts by role and dataset. Features shown are post–feature-selection.}
\label{app:tab:prompts}
\end{table*}

\section{Inference Efficiency}
\label{sec:efficiency}

We report additional measurements on (i) prompt length (input tokens), (ii) output length (generated tokens), and (iii) latency. We run inference with \textit{vLLM} and set the batch size to 1 to simulate an online scenario. For each dataset-model configuration, we measure LLM generation latency (in milliseconds) averaged over the evaluated queries. We also report the average number of input tokens and output tokens. For reasoning-capable models, we enable Thinking Mode for Qwen and set reasoning effort to Medium for GPT-OSS to reflect practical usage.

Table~\ref{tab:input_tokens} reports the average input-token count with and without FinFRE-RAG. As expected, FinFRE-RAG increases the prompt length substantially because it injects retrieved in-context exemplars and their labels.  Table~\ref{tab:lat_outtok} reports the average LLM generation latency together with the average number of output tokens. We count only the time spent on LLM generation, excluding the retrieval operation, as retrieval latency is highly system-dependent and can vary substantially with engineering choices (e.g., indexing, approximate nearest-neighbor search, and caching). In some settings, adding retrieved exemplars reduces average latency, as the presence of label-aware analogs enables the model to converge more quickly during generation and produce shorter responses. Conversely, for other configurations (e.g., GPT-OSS-20B on \textsc{ccFraud}), FinFRE-RAG increases both latency and output length, reflecting more detailed rationales when the model is conditioned on multiple exemplars.

\begin{table*}[t]
\centering
\setlength{\tabcolsep}{5pt}
\resizebox{0.9  \textwidth}{!}{
\begin{tabular}{lcc|cc|cc|cc}
\toprule
\multirow{2}{*}{\textbf{Model}} &
\multicolumn{2}{c|}{\textbf{\textsc{ccf}}} &
\multicolumn{2}{c|}{\textbf{\textsc{ccFraud}}} &
\multicolumn{2}{c|}{\textbf{\textsc{PaySim}}} &
\multicolumn{2}{c}{\textbf{\textsc{IEEE-CIS}}} \\
\cmidrule(lr){2-3}\cmidrule(lr){4-5}\cmidrule(lr){6-7}\cmidrule(lr){8-9}
& \#Tok w/o & \#Tok w/ 
& \#Tok w/o & \#Tok w/
& \#Tok w/o & \#Tok w/
& \#Tok w/o & \#Tok w/ \\
\midrule
Qwen3   & 227.01 & 2925.23 & 161.28 & 1543.06 & 212.13 & 2610.96 & 210.87 & 2161.63 \\
Gemma 3 & 224.01 & 2963.23 & 158.28 & 1581.06 & 206.71 & 2598.09 & 209.71 & 2225.74 \\
GPT-OSS & 262.00 & 2428.97 & 218.69 & 1518.26 & 243.70 & 2044.17 & 245.62 & 1756.99 \\
\bottomrule
\end{tabular}}
\caption{Average input-token counts (\#Tok) with and without FinFRE-RAG.}
\label{tab:input_tokens}
\end{table*}

\begin{table*}[t]
\centering
\setlength{\tabcolsep}{5pt}
\resizebox{\textwidth}{!}{
\begin{tabular}{lcc|cc|cc|cc}
\toprule
\multirow{2}{*}{\textbf{Model}} &
\multicolumn{2}{c|}{\textbf{\textsc{ccf}}} &
\multicolumn{2}{c|}{\textbf{\textsc{ccFraud}}} &
\multicolumn{2}{c|}{\textbf{\textsc{PaySim}}} &
\multicolumn{2}{c}{\textbf{\textsc{IEEE-CIS}}} \\
\cmidrule(lr){2-3}\cmidrule(lr){4-5}\cmidrule(lr){6-7}\cmidrule(lr){8-9}
& Latency (ms) & OutTok &
Latency (ms) & OutTok &
Latency (ms) & OutTok &
Latency (ms) & OutTok \\
\midrule
Qwen3-14B              & 18237.17 & 999.41 & 11057.95 & 540.26 &  9774.29 & 477.72 &  7748.80 & 378.37 \\
Qwen3-14B + FinFRE-RAG  & 13838.09 & 621.09 & 11855.37 & 560.6 & 12904.11 & 601.37 & 10637.88 & 498.24 \\
\midrule
Qwen3-Next-80B          & 31740.25 & 3756.54 & 15765.56 & 1855.07 & 15495.72 & 1822.56 & 45496.92 & 5368.77 \\
Qwen3-Next-80B + FinFRE-RAG
                        & 16472.47 & 1929.59 & 12608.28 & 1475.73 & 14964.63 & 1752.94 & 19905.43 & 2335.69 \\
\midrule
Gemma 3-12B            &  1415.67 & 73.47 &  1059.79 & 55.31 &  1452.68 & 75.55 &  1289.62 & 66.96 \\
Gemma 3-12B + FinFRE-RAG&  1300.47 & 49.42 &  1162.80 & 49.49 &  1387.43 & 55.76 &  1391.76 & 57.81 \\
\midrule
Gemma 3-27B            &  1592.12 & 101.91 &  1150.08 & 73.98 &  1383.32 & 88.56 &  1613.91 & 103.42 \\
Gemma 3-27B + FinFRE-RAG&  1390.03 & 70.65 &  1245.44 & 67.53 &  1583.24 & 87.61 &  1539.07 & 82.87 \\
\midrule
GPT-OSS-20B            &  1488.40 & 224.79 &  1638.03 & 248.97 &  1663.86 & 252.63 &  1569.56 & 235.72 \\
GPT-OSS-20B + FinFRE-RAG&  1600.38 & 221.47 &  2637.76 & 384.31 &  1398.36 & 193.96 &  2054.62 & 298.16 \\
\midrule
GPT-OSS-120B           &  1835.35 & 202.06 &  1928.25 & 213.78 &  2064.92 & 218.29 &  2384.15 & 249.35 \\
GPT-OSS-120B + FinFRE-RAG
                        &  1828.55 & 188.2 &  1962.71 & 209.95 &  1877.77 & 197.47 &  2207.12 & 247.35 \\
\bottomrule
\end{tabular}}
\caption{Average latency and average output tokens (OutTok).}
\label{tab:lat_outtok}
\end{table*}

\end{document}